\documentclass{article}
\usepackage{adjustbox}

    \usepackage[final,nonatbib]{neurips_2021_ml4ad}

\usepackage{amsfonts}       
\usepackage{amsmath}        
\usepackage{bbm}
\usepackage{booktabs}       
\usepackage[labelfont=bf,labelsep=period,font=small,hypcap=false]{caption}
\usepackage[T1]{fontenc}    
\usepackage{graphicx}       
\usepackage{hyperref}       
\usepackage[utf8]{inputenc} 
\usepackage{microtype}      
\usepackage{nicefrac}       
\usepackage{pgfplots}       
  \pgfplotsset{compat=1.9}
\usepackage{url}            
\usepackage{xcolor}         
\usepackage{xspace}         
\newcommand{\scenic}{{\sc Scenic}\xspace}
\newcommand{\verifai}{{\sc VerifAI}\xspace}

\title{A Scenario-Based Platform for Testing Autonomous Vehicle Behavior Prediction Models in Simulation}

\author{
    Francis Indaheng$^1$ \\
        \texttt{findaheng@berkeley.edu}
    \And Edward Kim$^1$ \\
        \texttt{ek65@eecs.berkeley.edu}
    \And Kesav Viswanadha$^1$ \\
        \texttt{kesav@berkeley.edu}
    \And Jay Shenoy$^1$ \\
        \texttt{jayshenoy@berkeley.edu}
    \And Jinkyu Kim$^2$ \\
        \texttt{jinkyukim@korea.ac.kr}
    \And Daniel J. Fremont$^3$ \\
        \texttt{dfremont@ucsc.edu}
    \And Sanjit A. Seshia$^1$ \\
        \texttt{sseshia@eecs.berkeley.edu} \\
    \and
        $^1$University of California, Berkeley \\
        $^2$Korea University \\
        $^3$University of California, Santa Cruz
}

\begin{document}

\maketitle

\begin{center}
    \includegraphics[scale=0.4]{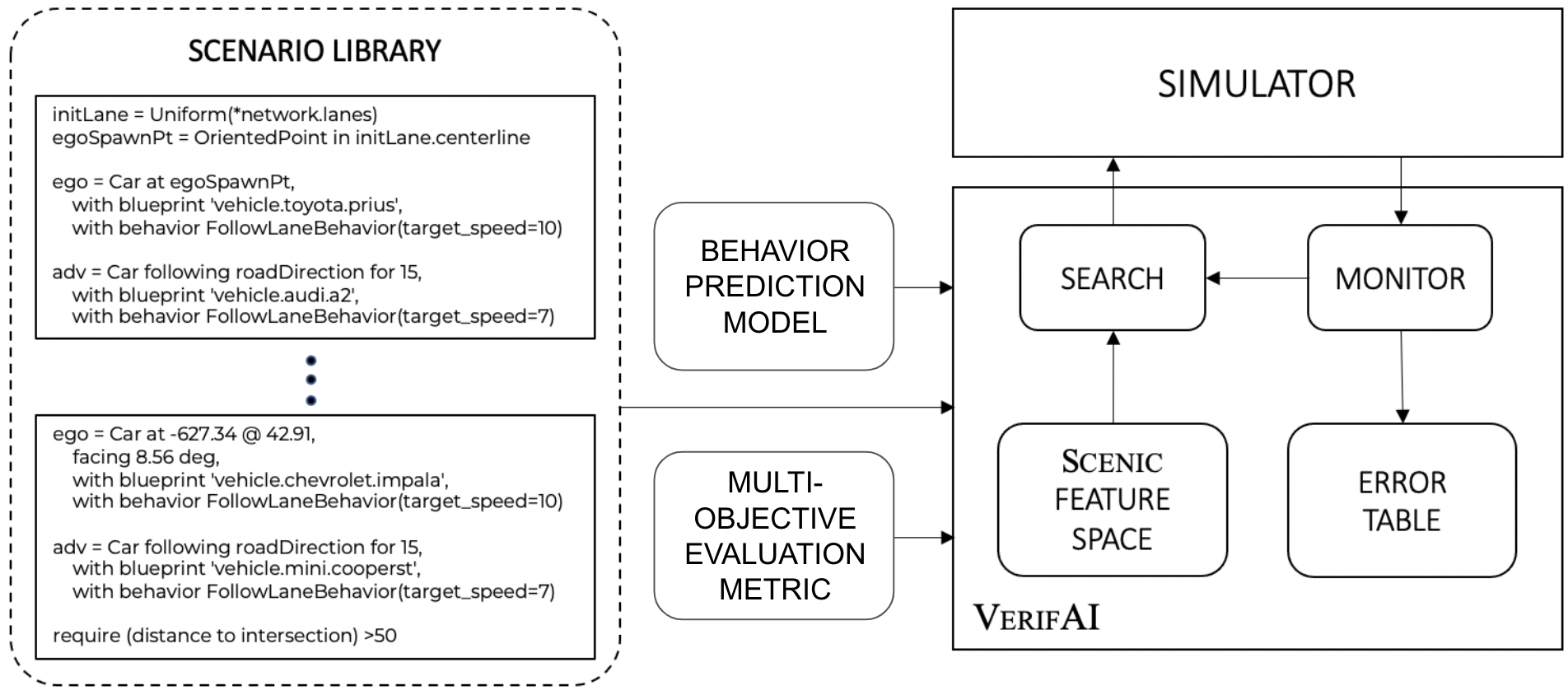}
    \captionof{figure}{An architecture diagram demonstrating the flow of components in the testing platform. A behavior prediction model, library of \scenic test scenarios, and multi-objective evaluation metric serve as inputs to \verifai, which monitors and searches for failure cases.}
    \label{fig:architecture}
    \vspace{5mm}
\end{center}

\begin{abstract}
Behavior prediction remains one of the most challenging tasks in the autonomous vehicle (AV) software stack. Forecasting the future trajectories of nearby agents plays a critical role in ensuring road safety, as it equips AVs with the necessary information to plan safe routes of travel. However, these prediction models are data-driven and trained on data collected in real life that may not represent the full range of scenarios an AV can encounter. Hence, it is important that these prediction models are extensively tested in various test scenarios involving interactive behaviors prior to deployment. To support this need, we present a simulation-based testing platform that supports (1) intuitive scenario modeling with a probabilistic programming language called \scenic, (2) specifying a multi-objective evaluation metric with a partial priority ordering, (3) falsification of the provided metric, and (4) parallelization of simulations for scalable testing. As a part of the platform, we provide a library of 25 \scenic programs that model challenging test scenarios involving interactive traffic participant behaviors. We demonstrate the effectiveness and the scalability of our platform by testing a trained behavior prediction model and searching for failure scenarios.

\end{abstract}

\section{Introduction}
Behavior (or trajectory) prediction, which is a component of an autopilot, plays a critical role in ensuring road safety, as an autonomous vehicle (AV) that can accurately forecast the motions of surrounding vehicles is better equipped to plan safe routes of travel. This prediction task is challenging and is currently an active field of study. The difficulty of this task stems from the multifaceted nature of dynamic traffic scenarios: an effective model must appropriately account for both the AV’s surrounding environment and the social interactions between various agents over a duration of time.

Models for such tasks are commonly based on neural network (NN) architectures, such as Recurrent Neural Networks (RNNs)~\cite{Homayounfar_2018_CVPR} and Long Short-Term Memory (LSTM) networks~\cite{10.1162/neco.1997.9.8.1735, DBLP:journals/corr/abs-1910-03650}. These are data-driven models trained on data collected on roads. Capturing data related to all possible scenarios on roads is difficult, let alone characterizing all possible scenarios itself. There are edge-case (i.e. rare) scenarios that, by definition, are rarely encountered on roads and, consequently, are only sparsely reflected in the training data, if at all. Hence, testing behavior prediction models in, at least, \textit{known} edge case scenarios is crucial. Tending to this issue in the real world would involve reconstructing such scenarios in a controlled testing facility, which can be expensive and labor-intensive, resulting in a lack of scalability of testing. On the contrary, simulation offers an inexpensive mechanism for reconstructing scenarios of interest, making for a more scalable approach. Yet, simulation-based testing of behavior prediction models requires  support for (1) modeling interactive scenarios of interest in simulation, (2) generating necessary input synthetic sensor data from those simulations, and (3) evaluating behavior prediction models using multi-objective evaluation metrics. We aim to address this need in this paper.

\paragraph{Contributions}

We provide a platform to test behavior prediction models that is efficient, scalable, and scenario-based. The main contributions of this work include the following:

\begin{itemize}
    \item An open-source platform\footnote{\href{https://github.com/BehaviorPredictionTestingPlatform/TestingPipeline}{https://github.com/BehaviorPredictionTestingPlatform/TestingPipeline}} for researchers and developers to evaluate behavior prediction models in the CARLA simulator~\cite{Dosovitskiy17} under scenarios modeled in the \scenic language~\cite{Fremont_2019, fremont2020scenic}. The platform allows users to specify custom multi-objective evaluation metrics and falsify their models with respect to these metrics, parallelizing simulations for scalable testing;
    \item An open-source library of 25 \scenic programs to evaluate behavior prediction models in various dynamic scenarios;\footnote{\href{https://github.com/BehaviorPredictionTestingPlatform/ScenarioLibrary}{https://github.com/BehaviorPredictionTestingPlatform/ScenarioLibrary}}
    \item An experimental evaluation of a behavior prediction model against five traffic scenarios to demonstrate the efficacy and scalability of our platform.
\end{itemize}

\section{Background}\label{sec:background}
\subsection{\scenic: Scenario Modeling Language}\label{sec:scenic}

\scenic~\cite{Fremont_2019,fremont2020scenic} is a probabilistic programming language whose syntax and semantics are designed to intuitively \textit{model} and \textit{generate} scenarios involving interactive agents. A \scenic program represents an \textit{abstract} scenario which defines distributions over the initial states and dynamic behaviors of agents in the scenario. Sampling from these distributions yields a \textit{concrete} scenario that assigns specific values to these parameters.
The \scenic tool can generate concrete scenarios from a \scenic program and execute them in a simulator to generate data and test a system.
\scenic is simulator-agnostic and has interfaces to a variety of simulators for driving, aviation, robotics, and other domains~\cite{fremont2020scenic}.

\subsection{\verifai: Formal Analysis Software Toolkit}\label{sec:verifai}

\verifai~\cite{verifai-cav19, viswanadha_2021} is a toolkit for formal design and analysis of AI/ML-based systems. Its architecture is shown in Figure~\ref{fig:architecture}. It takes three inputs: (1) the system of interest (e.g. behavior prediction model), (2) an environment model (e.g. \scenic program), and (3) a multi-objective evaluation metric. 

\verifai performs falsification by sampling concrete scenarios using a variety of techniques, simulating each one and monitoring the system’s performance with respect to the given evaluation metric. The sampled test parameters and corresponding system performance are logged for further analysis, and also provided to the sampling algorithms to more intelligently select the next scenario. To support multi-objective evaluation metrics, \verifai provides a multi-armed bandit sampler, which formulates the search process as a multi-armed bandit problem with a reward function favoring concrete scenarios that pessimize multiple metrics (taking priority into account)~\cite{viswanadha_2021}.
\section{Overview}\label{sec:overview}
The overall architecture of our platform is visualized in Figure~\ref{fig:architecture}. The user needs to provide three inputs: (1) a trained behavior prediction model, (2) a multi-objective evaluation metric, and (3) a library of \scenic programs modeling a set of abstract scenarios (refer to Section~\ref{sec:scenic}).

The evaluation pipeline for behavior prediction is shown in Figure~\ref{fig:multi_obj} in the Appendix. As stated in Section~\ref{sec:scenic}, a \scenic program in conjunction with the CARLA simulator is used to generate various concrete scenarios. We implement support for collecting various types of data from each simulation to enable testing different behavior prediction models, which can take various types of inputs. Users can collect trajectories (i.e. timestamped position and headings of all traffic participants in the scenario) and a stream of sensor data from RGB camera(s), depth camera(s), and/or LiDAR sensor(s). We can also collect corresponding ground-truth labels such as segmentation labels for RGB images and LiDAR, as well as 3D bounding boxes. 

From these collected data, we extract two different types of data: (1) historical data, which is the segment of the data which is input to the behavior prediction model, and (2) ground-truth future trajectory data, which is used to evaluate the model's predicted trajectories using the provided multi-objective evaluation metric, which outputs $\vec{\rho}$, a tuple containing a \textit{score} for each metric specified. \verifai's multi-armed bandit sampler is used to falsify the given multi-objective evaluation metric. Please refer to Section~\ref{sec:verifai} for more detail on the sampler. 

\section{Experimental Evaluation}
\begin{table*}
\centering
\begin{tabular}{ c||ccc|cc }
    \toprule
    Scenario \# & minADE & minFDE & MR & CR & SD \\
    \hline
    \rule{0pt}{1em}\ref{scenario_1} & 0.037 & 0.15 & 0.017 & 0.034 & 0.58 \\
    \ref{scenario_2} & 0.044 & 0.19 & 0.016 & 0.025 & 0.54 \\
    \ref{scenario_3} & 0.081 & 0.39 & 0.034 & 0.24 & 0.45 \\
    \ref{scenario_4} & 0.13 & 0.74 & 0.31 & 0.36 & 0.52 \\
    \ref{scenario_5} & 0.19 & 1.1 & 0.49 & 0.51 & 0.56 \\
    \bottomrule
\end{tabular}
\captionof{table}{Results of evaluating the LaneGCN model against five \scenic programs. The minADE, minFDE, and MR metrics indicate the model's error of predictions with respect to ground-truth trajectories. The CR and SD metrics reflect the ability of the \verifai sampler to identify a diverse number of counterexample scenarios.}
\label{tab:metrics}
\end{table*}


We used our platform to evaluate the LaneGCN model~\cite{liang2020learning}; for details of our setup, see Appendix~\ref{sec:exp_setting}.

In Table~\ref{tab:metrics}, we enumerate the results from our testing platform after evaluating LaneGCN against the five scenarios in Appendix~\ref{sec:scenarios}. The first three metrics (minADE, minFDE, MR) quantify the behavior prediction model's performance, averaged over all samples generated by \verifai, whereas the latter two metrics, counterexample rate (CR) and scenario diversity (SD), provide insight about the effectiveness of different sampling methods. The CR expresses the number of counterexamples found in \verifai with respect to the total number of samples run. The SD relates the observed variance to the defined range of values for a set of feature distributions. For further details on these metrics, refer to Appendix~\ref{sec:metrics}.

While SD remained relatively stable around 0.53$\pm$0.08, we observed some variance in the other metrics. LaneGCN made noticeably more accurate predictions for Scenarios~\ref{scenario_1},~\ref{scenario_2}, and~\ref{scenario_3} than for Scenarios~\ref{scenario_4} and~\ref{scenario_5}, reflected by their substantially lower minADE, minFDE, and MR values. We speculate this may be attributed to the differences in complexity of the vehicle's behavior in each of these scenarios. In Scenarios~\ref{scenario_1},~\ref{scenario_2}, and~\ref{scenario_3}, the oncoming vehicles had a relatively straight trajectory, either continuing a straight path or making a single lane change. Meanwhile, in Scenarios~\ref{scenario_4} and~\ref{scenario_5}, the oncoming vehicles makes an unprotected left turn across an intersection. Scenario~\ref{scenario_5} is even more complex than Scenario~\ref{scenario_4}, with the ego vehicle making a right turn at the intersection instead of continuing straight. Recall that LaneGCN takes historical trajectories for \textit{all} traffic agents as input, hence why the ego vehicle's own behavior could influence the accuracy of its predictions. Overall, these results suggest that our scenario-based testing platform is effective for evaluating behavior prediction model performance and identifying difficult driving scenarios for the model.


To demonstrate the efficiency of our approach, we benchmarked the runtime of the testing platform for both sequential and parallel executions. We conducted these experiments by evaluating LaneGCN against Scenario~\ref{scenario_1} for N iterations, where $N\in\{25, 50, 75, 100\}$. We ran the pipeline sequentially, during which there was no overlap between sampling and simulation, and then we ran the pipeline in parallel, first with 2 workers, then with 5 workers.

The results are plotted in Figure~\ref{fig:runtime} in the Appendix. On average, parallel execution yielded a 1.51x speedup with 2 workers and a 3.09x speedup with 5 workers. There are a number of reasons the speedup did not quite achieve a factor consistent with the number of parallel workers. One definite cause is that \verifai only parallelizes simulation and evaluation~\cite{viswanadha_2021}. Thus, while simulation of the concrete scenarios can occur simultaneously on several instances of the CARLA simulator, the sampling process remains sequential.

\pgfplotstableread{
iter   w1    w2    w5
 25 12.95  8.17  4.12
 50 23.32 16.33  7.98
 75 36.92 24.00 11.47
100 47.35 31.80 15.30
}\runtimedata

\section{Related Work}
Although there have been increasing efforts to develop behavior prediction models for AVs, there has been little work on formally testing them in simulation. Some work has evaluated AV planners using simulated perception and prediction inputs~\cite{wong2020testing} from a set of scenarios, but the focus was to demonstrate applying different noise models to improve the realism of simulated data. Our efforts are directed towards creating an efficient and scalable testing platform that provides customizable metrics and data generation capabilities, with the capacity for users to model their own scenarios.

\verifai has been used in a number of other case studies on AI systems~\cite{verifai-cav19}, including: identifying relevant tests for AV track testing~\cite{fremont2020formal}; falsifying, debugging, and retraining an autonomous aircraft taxiing system~\cite{fremont2020aircraft}; and evaluating a full autopilot stack in the IEEE AV Test Challenge~\cite{viswanadha2021addressing}.
Our work is the first to use \verifai for testing behavior prediction models.

\section{Conclusion}
In this paper, we presented a platform for testing behavior prediction models with \scenic and \verifai. As part of the platform, we provided a library of interactive scenarios encoded as \scenic programs. We demonstrated our platform by evaluating the LaneGCN model against well-known motion forecasting metrics. Lastly, we showcased the scalability of running the pipeline in a parallelized fashion with a runtime benchmark comparison to a sequential execution.

\begin{ack}
The authors would like to express thanks to Hazem Torfah, Sebastian Junges, Marcell Vazquez-Chanlatte, Yash Pant, Justin Wong, and Ameesh Shah for their helpful discussions and feedback. This work is supported in part by NSF grants 1545126 (VeHICaL) and 1837132, by the DARPA contracts FA8750-18-C-0101 (AA) and FA8750-20-C-0156 (SDCPS), by Berkeley Deep Drive, by the Toyota Research Institute, and by Toyota under the iCyPhy center.

\end{ack}

\bibliographystyle{plain}

\vfill

\pagebreak
\appendix
\section{Description of the Library of \scenic Programs}\label{sec:scenic_library}
As a part of our behavior prediction evaluation platform, we provide a library of 25 \scenic programs as test scenario models. We referred to the INTERACTION dataset~\cite{interactiondataset}, the Argoverse Motion Forecasting dataset~\cite{Chang_2019_CVPR}, and a publication by the National Highway Traffic Safety Administration (NHTSA) of pre-crash scenarios based on 2011-2015 crash data recorded by the U.S. Department of Transportation~\cite{nhtsa2019} and encoded scenarios as \scenic programs. These datasets were chosen based on our efforts to capture a diverse selection of \textit{interactive} scenarios with \textit{critical} trajectories (e.g., near-collision situations). These scenarios covered a wide variety of road structures—such as highways, intersections, and roundabouts—as well as agent interactions—such as bypassing, merging, and unprotected left turns. 

\section{Metrics}\label{sec:metrics}
\subsection{Multi-Objective Evaluation Metrics}\label{section:eval_metrics}

In \verifai, a multi-objective function defines a set of metrics used to determine the efficacy of the system being evaluated. As stated in Section~\ref{sec:overview}, the output of the function is \textit{$\vec{\rho}$}, a tuple containing a \textit{score} for each metric specified, the calculation of which is defined in the function. By convention, a negative score for any element in $\vec{\rho}$ indicates a failure, in which case the counterexample is stored in the error table for offline analysis; accordingly, a non-negative score for all elements in $\vec{\rho}$ indicates the model succeeded in meeting the evaluation criteria.

For the purpose of testing behavior prediction models, we apply three commonly-used motion forecasting metrics. \textit{Average Displacement Error} (ADE) is the Euclidean distance between the predicted trajectories, $(\hat{x}_t, \hat{y}_t)$, and the ground-truth trajectories, $(x_t, y_t)$, averaged over all $n$ timesteps:

\begin{equation}
    ADE(x,y)=\frac{1}{n}\sum_{t=1}^n\sqrt{(x_t-\hat{x}_t)^2+(y_t-\hat{y}_t)^2}
\end{equation}

\textit{Final Displacement Error} (FDE) is the Euclidean distance between the predicted trajectory, $(\hat{x}_n, \hat{y}_n)$, and the ground-truth trajectory, $(x_n, y_n)$, at the last timestep:

\begin{equation}
    FDE(x,y)=\sqrt{(x_{n}-\hat{x}_{n})^2+(y_{n}-\hat{y}_{n})^2}
\end{equation}

For the $i^{th}$ sample in a dataset, we take the minimum ADE (minADE) and minimum FDE (minFDE) across the $k$ most probable predictions, $P_k^i$, generated by the model, as done in similar approaches accommodating multi-modal trajectories~\cite{cui2019multimodal, khandelwal2020if, liang2020learning, DBLP:journals/corr/abs-1910-03650}. Doing so avoids penalizing the model for producing plausible predictions that don’t align with the ground truth.

\begin{align}
    minADE(P_k^i) &= min_{(x,y)\in P_k^i}ADE(x,y)\label{eq:minADE} \\
    minFDE(P_k^i) &= min_{(x,y)\in P_k^i}FDE(x,y)\label{eq:minFDE}
\end{align}

The third metric, \textit{Miss Rate} (MR), is the ratio of missed predictions over the entire test set of $N$ data points, $P_k$. In this context, a missed prediction occurs when all $k$ proposed trajectories in prediction $P_k^i$ differ in final position from the ground truth by more than some distance $d$ threshold. This metric indicates the trade-off between accuracy and diversity~\cite{mercat2019multihead}.

\begin{equation}
    MR(P_k,d)=\frac{1}{N}\sum_{i=1}^{N}\mathbbm{1}\{minFDE(P_k^i)>d\}\label{eq:MR}
\end{equation}

In our platform, we enabled a command-line argument for the user to specify the threshold values for minADE, minFDE, and MR. Recall that $\vec{\rho}$ is a tuple of scores, and if any score in $\vec{\rho}$ is negative, then the scenario is deemed a failing counterexample. To conform to this convention, we store the difference between the thresholds and their corresponding metrics as the scores in $\vec{\rho}$ (e.g., we store $(2-minADE)$ to enforce a minADE threshold of 2).
\subsection{Platform Metrics}\label{sec:platform_metrics}

In section \ref{section:eval_metrics}, we discussed the metrics that would be used to evaluate the behavior prediction model (Eq. \ref{eq:minADE}, \ref{eq:minFDE}, \ref{eq:MR}). Now, we describe two metrics for evaluating the effectiveness of our testing platform: \textit{Counterexample Rate} (CR) and \textit{Scenario Diversity} (SD). The CR expresses the number of counterexamples found in \verifai with respect to the total number of samples run. Thus, the CR reflects the ability of the multi-armed bandit sampler to search the feature space and identify failing scenarios in which the behavior prediction model's predictions did not meet the criteria of the multi-objective function:

\begin{equation}
    CR=\frac{\textnormal{Number of counterexamples}}{\textnormal{Number of all examples}}
\end{equation}

We can also formalize the overall SD produced by \scenic, which relates the observed variance to the defined range of values for a set of feature distributions. For the $i^{th}$ feature, $f_i$, let $\sigma(f_i)$ denote the standard deviation of values observed and $L(f_i)$ denote the interval length, where the interval length is the size of the range of values specified in the feature's distribution. The SD for a set of $N$ features is as follows:

\begin{equation}
    SD=\frac{2\cdot\sum_{i=1}^N\sigma(f_i)}{\sum_{i=1}^NL(f_i)}
\end{equation}

\section{LaneGCN Behavior Prediction Model}

We evaluate LaneGCN behavior prediction model by Uber ATG~\cite{liang2020learning}. This behavior prediction model ranked $1^{st}$ in a motion forecasting competition~\cite{argoAICompetition} held by Argo AI, based on the Argoverse dataset~\cite{Chang_2019_CVPR}. Furthermore, it was one of few working open-sourced behavior prediction models. We evaluate the pre-trained LaneGCN model, which was trained on over 300,000 real-life scenarios collected for the aforementioned Argoverse dataset.

To test the model in CARLA, we implemented a support to make it operational with the OpenDRIVE format~\cite{Dupuis2006OpenDRIVE, Dupuis2010OpenDRIVE2A}. We applied the necessary computations to bridge the gap between map representations in \scenic and the model. Due to the modular nature of the model's architecture, this functional exchange could be completed with minimal effect to any dependencies.

The model expects a CSV file containing 20 timesteps of historical trajectories for all traffic agents as input. It produces 6 predictions, each containing 15 timesteps of predicted trajectories, for a designated single traffic agent. All trajectories use a sample rate of 10 Hz.

\section{Experimental Details}
\subsection{Experimental Settings}\label{sec:exp_setting}

For these experiments, we utilized the multi-armed bandit sampler in \verifai for searching the feature space. As demonstrated in a recent case study~\cite{viswanadha_2021}, the multi-armed bandit sampler strikes a strong balance between falsification capabilities (i.e., the number of counterexamples found) and feature space diversity (i.e., the concrete values sampled for each feature). Please refer to Section~\ref{sec:background} for more detail. 

For each scenario, we sample and generate 120 concrete scenarios, divided into 4 batches. The difference between batches was the initial point in time we wished to start the prediction, referred to as the parameter \textit{timepoint}. For example, setting \texttt{timepoint=40} indicates historical trajectories encompass timesteps in the range [20, 40) and predicted trajectories encompass timesteps in the range [40, 55). Note that since historical trajectories include 20 timesteps, there is a minimum constraint of \texttt{timepoint=20}. This parameter allows users to test a model on different parts of a scenario, which may involve greater or lesser complexity of agent-to-agent interactions. Thus, we generated 30 samples (i.e., a batch) for each of \texttt{timepoint$\in$\{20,40,60,80\}} to evaluate the behavior prediction model's capability to predict in different segments of a scenario. We set the minADE threshold to 0.1 meters, the minFDE threshold to 1.0 meter, and the MR distance threshold to 1.0 meter.

These experiments were conducted on a machine running Ubuntu 18.04.5, equipped with an Intel Core i9-9900X @ 3.50GHz central processing unit (CPU), as well as an Nvidia GEForce GTX and two Nvidia Titan RTX graphics processing units (GPUs). 

\subsection{Scenario Descriptions}\label{sec:scenarios}

We illustrate use of our testing platform by evaluating the LaneGCN model against five scenarios of varying complexity from our library of \scenic programs, described as the following:

\begin{enumerate}
    \item The ego vehicle waits at a 3-way intersection for another vehicle from the lateral lane to pass before making a right turn.\label{scenario_1}
    \item The ego vehicle makes a left turn at a 3-way intersection and must maneuver to avoid collision when another vehicle from the lateral lane continues straight.\label{scenario_2}
    \item A trailing vehicle performs a lane change to bypass the leading ego vehicle before returning to its original lane.\label{scenario_3}  
    \item The ego vehicle drives straight through a 4-way intersection and must suddenly stop to avoid collision when another vehicle from the oncoming lane makes an unprotected left turn.\label{scenario_4}
    \item The ego vehicle makes a right turn at a 4-way intersection while another vehicle from the oncoming lane makes an unprotected left turn, such that both vehicles are turning to the same outgoing leg of an intersection.\label{scenario_5}
\end{enumerate}

Here, \textit{ego vehicle} refers to the reference point for the behavior prediction (i.e., the vehicle in the scenario that would be administering the behavior prediction model). The \scenic source code of these scenarios is included in the appendix as Figures~\ref{fig:intersection_09}-\ref{fig:intersection_05}.

\begin{center}
    \includegraphics[scale=0.5]{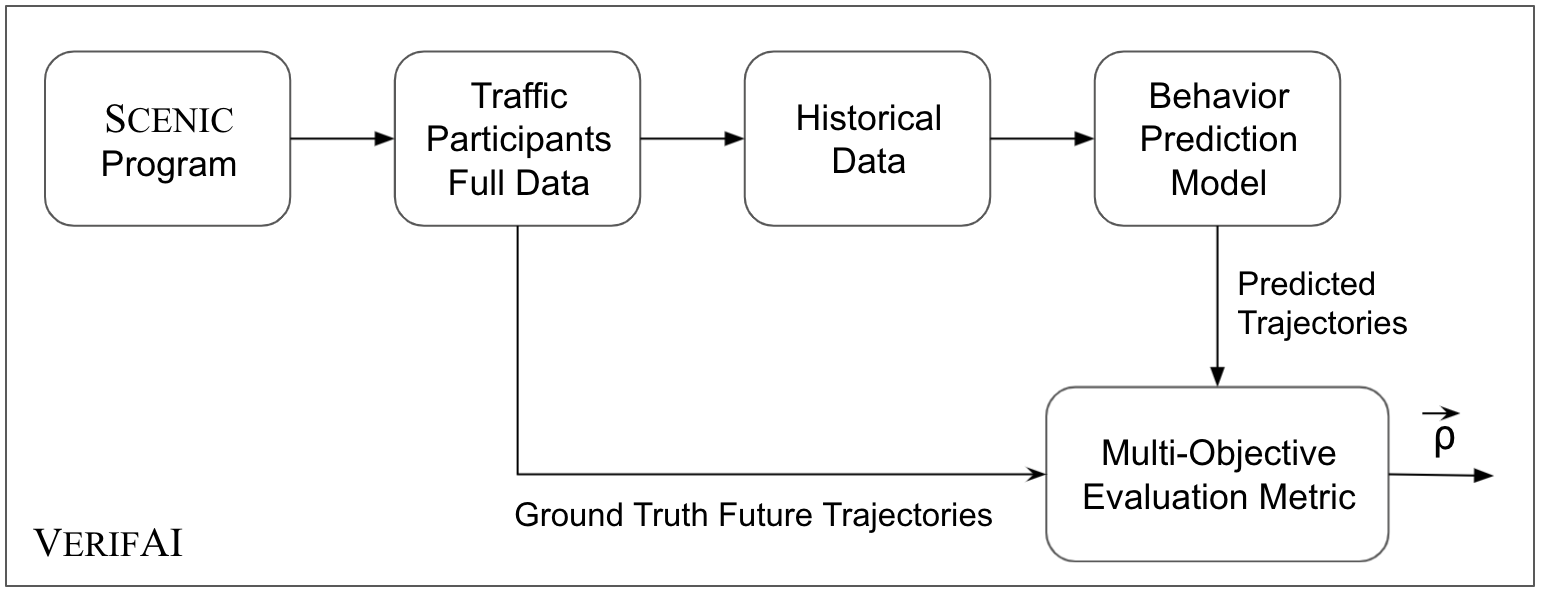}
    \captionof{figure}{An overview of the testing pipeline in \verifai. Historical data is extracted and serves as input to the behavior prediction model. The predicted and ground-truth trajectories are compared against the multi-objective evaluation metric to derive a tuple of scores, referred to as $\vec{\rho}$.}
    \label{fig:multi_obj}
\end{center}

\begin{figure}
    \centering
    \includegraphics[scale=0.35]{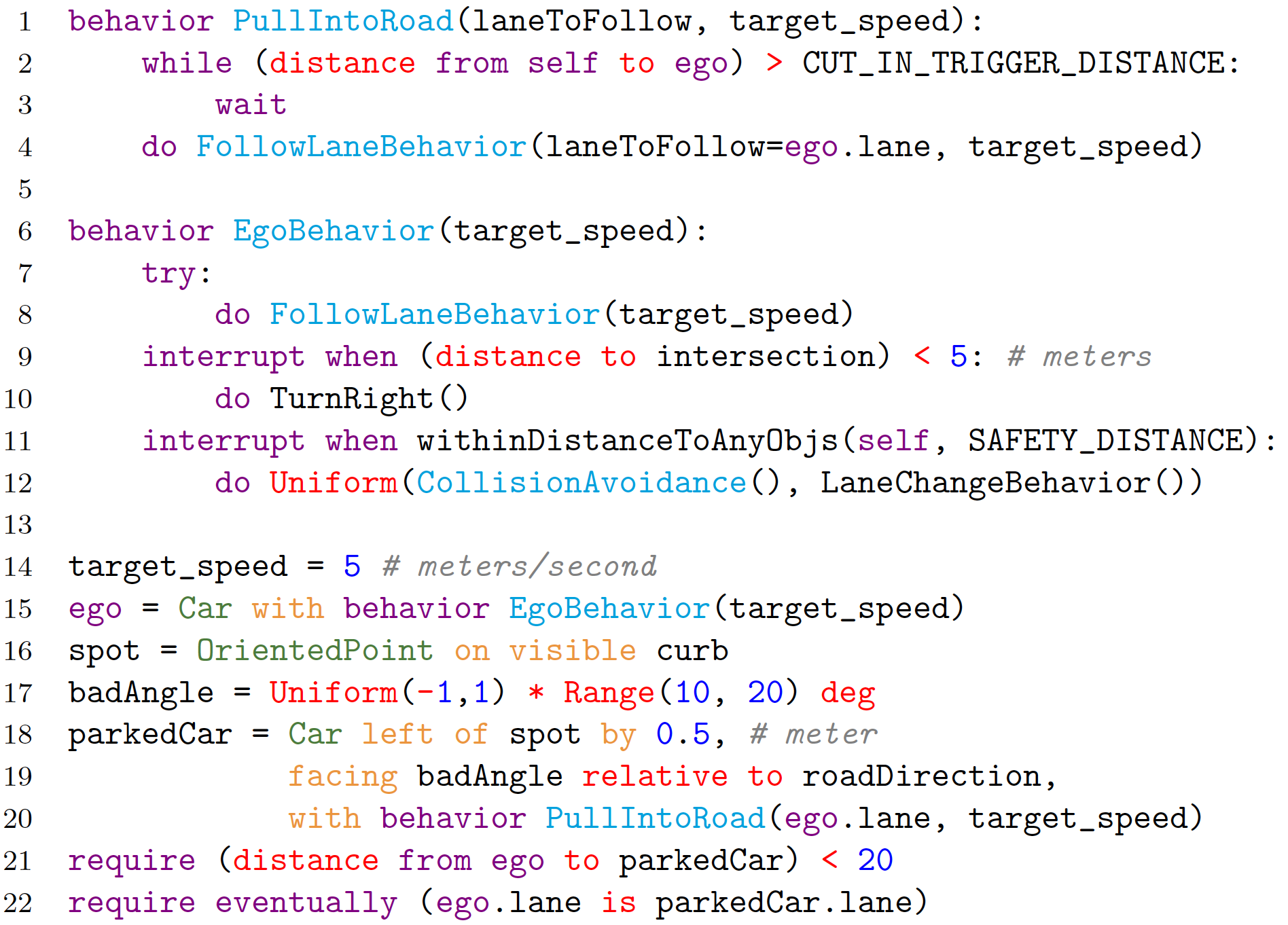}
    \caption{An example \scenic program in which a badly parked car cuts into the ego vehicle's lane.}
    \label{fig:scenic_program}
\end{figure}

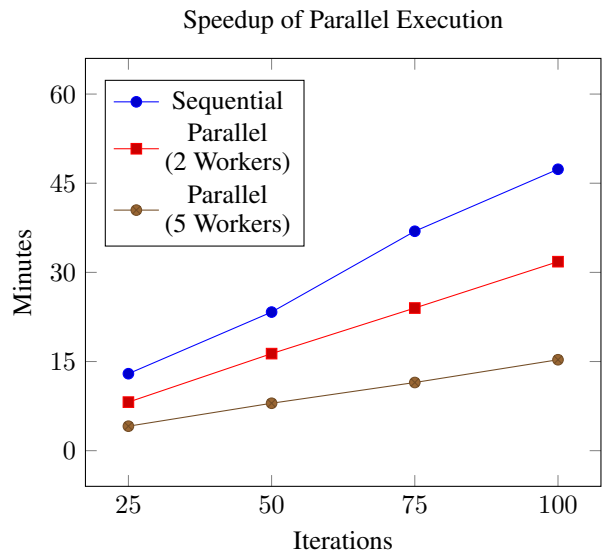
\begin{figure}
\centering
\begin{tikzpicture}
\begin{axis}[
	x tick label style={/pgf/number format/1000 sep={}},
	title=Speedup of Parallel Execution,
	xlabel=Iterations,
	ylabel=Minutes,
	xtick={25, 50, 75, 100},
	ytick={0, 15, 30, 45, 60},
	ymin=0, ymax=60,
	enlargelimits=0.1,
	legend style={
	    at={(0.425, 0.95)},
	    cells={align=center}
    }
]
\addplot table[x=iter,y=w1] {\runtimedata};
\addplot table[x=iter,y=w2] {\runtimedata};
\addplot table[x=iter,y=w5] {\runtimedata};
\legend{Sequential,Parallel\\(2 Workers),Parallel\\(5 Workers)}
\end{axis}
\end{tikzpicture}
\captionof{figure}{Runtime benchmark of sequential and parallel executions in the testing pipeline. Iterations is the number of simulations ran.}
\label{fig:runtime}
\end{figure}

\pagebreak

\begin{center}
\centering
\includegraphics[scale=0.64]{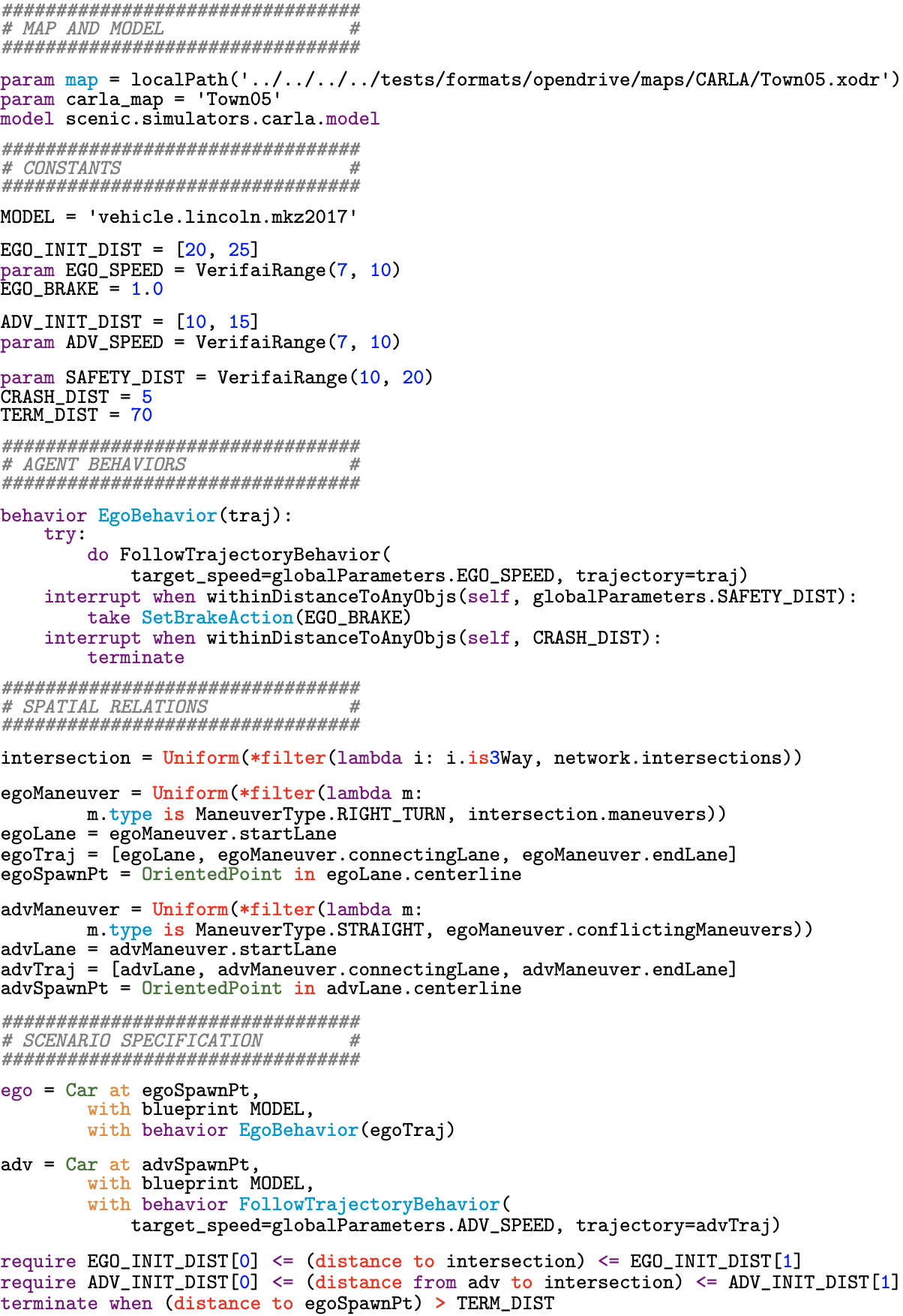}
\captionof{figure}{\scenic program used in Scenario \ref{scenario_1}}
\label{fig:intersection_09}
\end{center}

\pagebreak

\begin{center}
\centering
\includegraphics[scale=0.65]{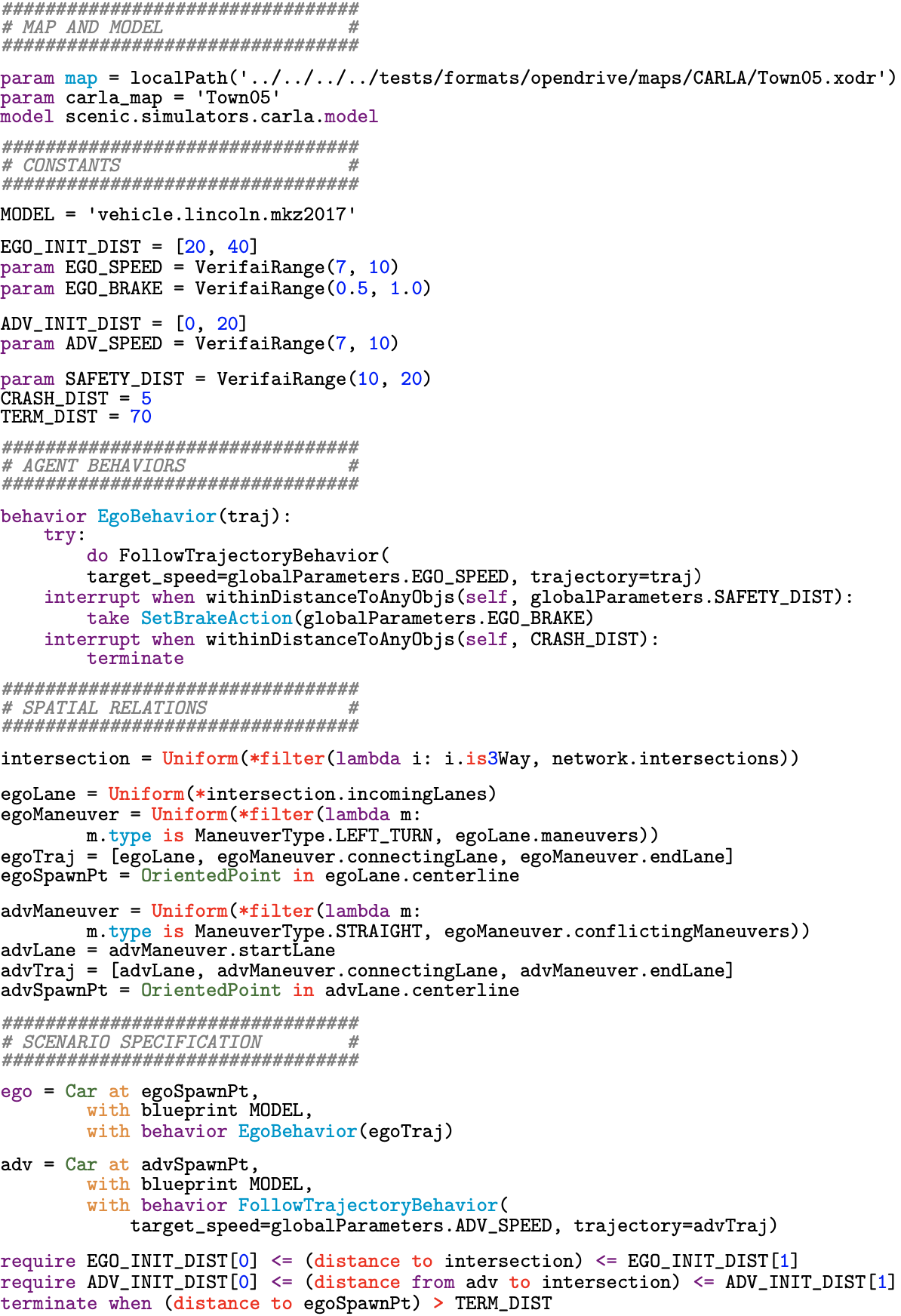}
\captionof{figure}{\scenic program used in Scenario \ref{scenario_2}}
\label{fig:intersection_07}
\end{center}

\pagebreak

\begin{center}
\centering
\includegraphics[scale=0.63]{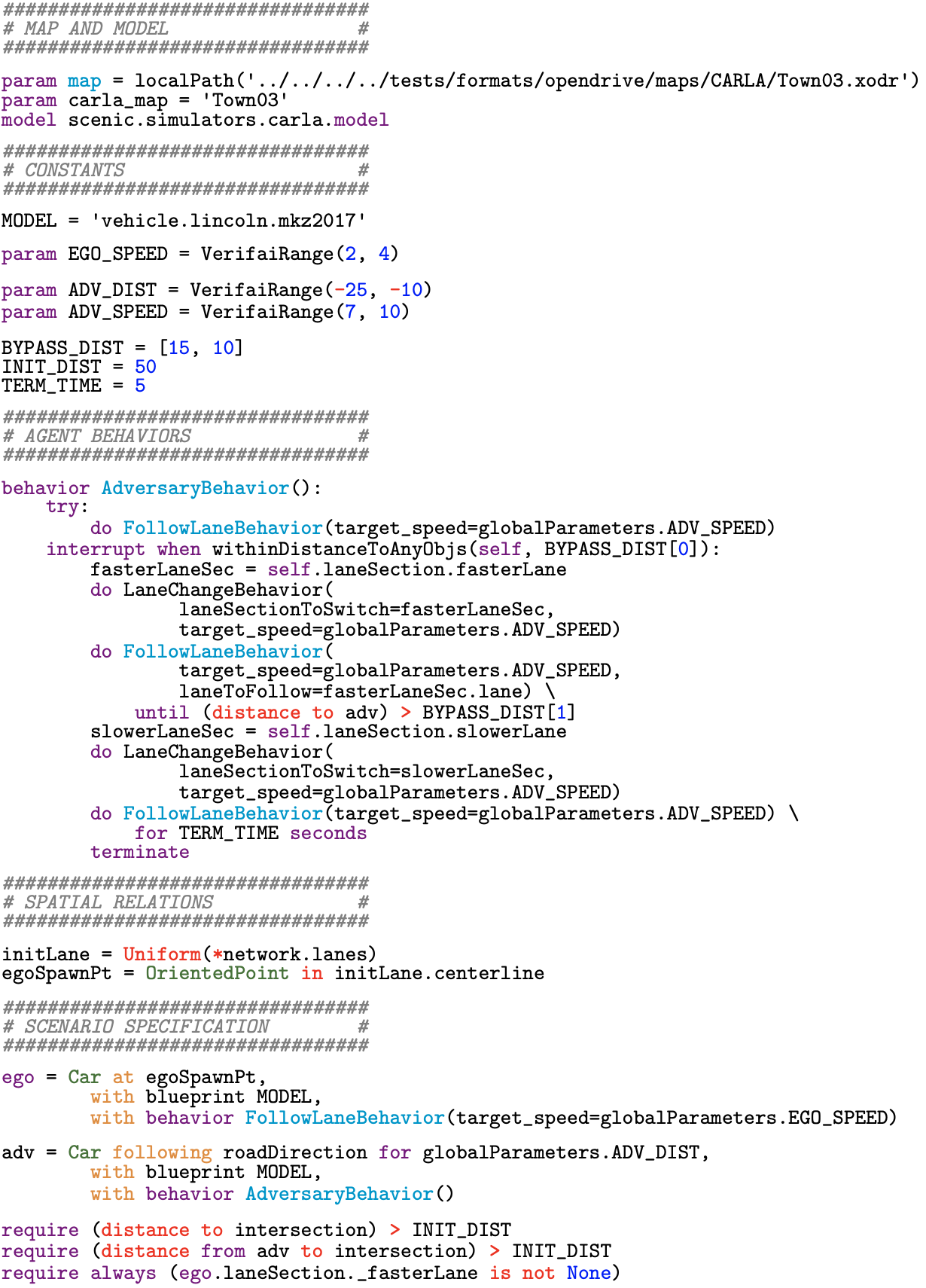}
\captionof{figure}{\scenic program used in Scenario \ref{scenario_3}}
\label{fig:bypassing_02}
\end{center}

\pagebreak

\begin{center}
\centering
\includegraphics[scale=0.67]{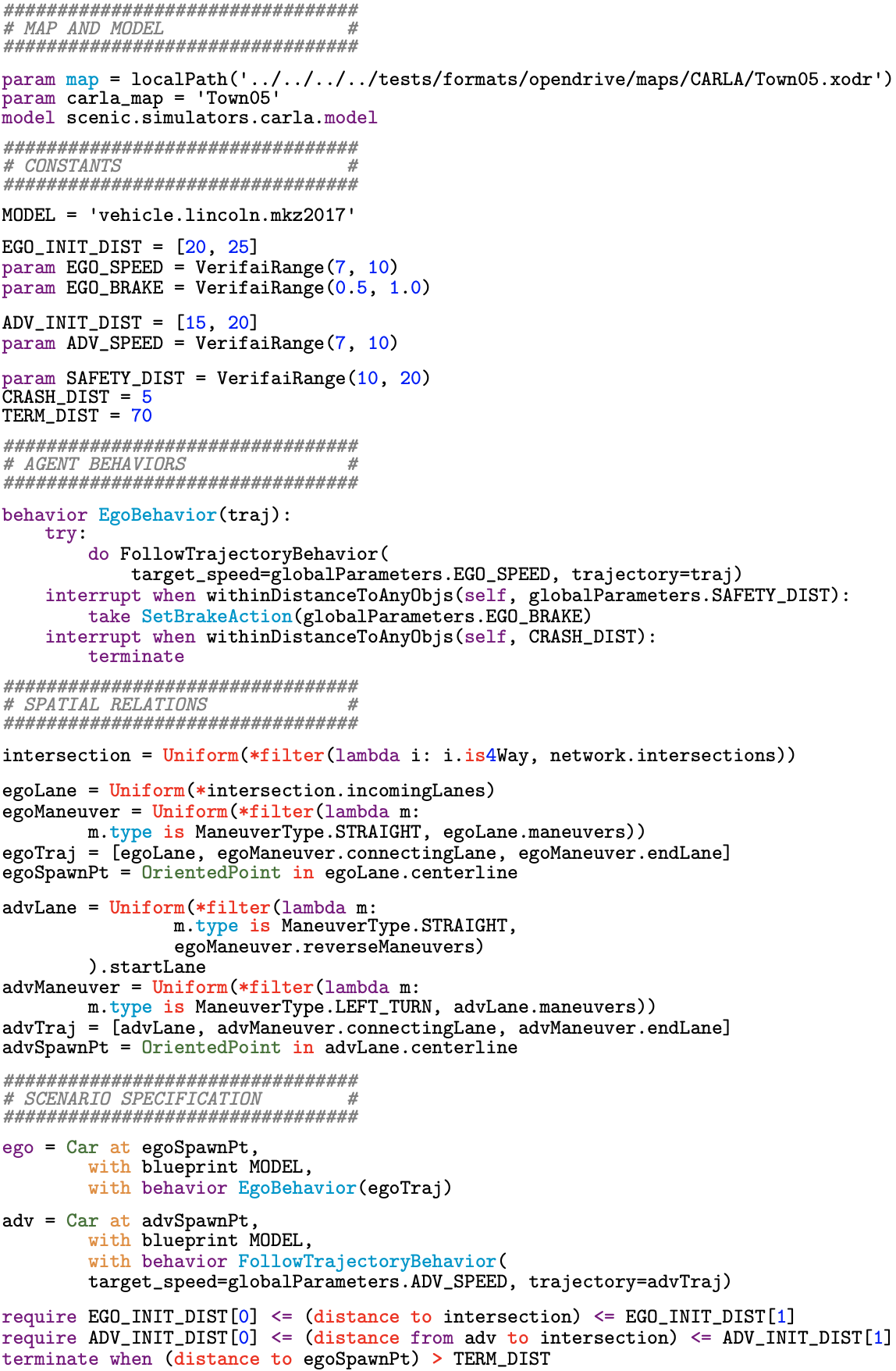}
\captionof{figure}{\scenic program used in Scenario \ref{scenario_4}}
\label{fig:intersection_01}
\end{center}

\pagebreak

\begin{center}
\centering
\includegraphics[scale=0.7]{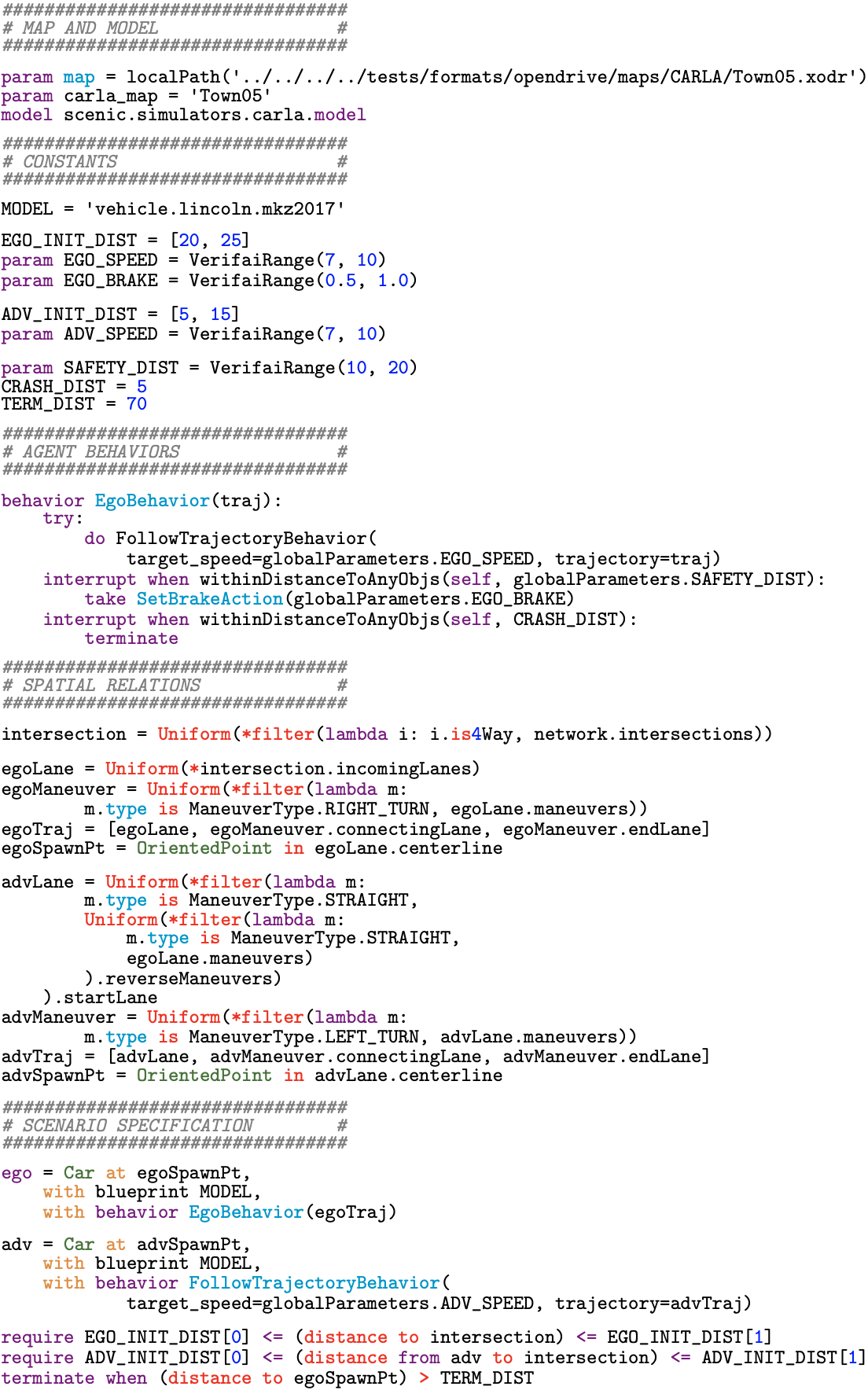}
\captionof{figure}{\scenic program used in Scenario \ref{scenario_5}}
\label{fig:intersection_05}
\end{center}

\end{document}